\documentclass[runningheads]{llncs}
\usepackage{graphicx}
\usepackage{comment}
\usepackage{amsmath,amssymb} 
\usepackage{color}

\usepackage[utf8]{inputenc} 
\usepackage{url}            
\usepackage{booktabs}       
\usepackage{amsfonts}       
\usepackage{nicefrac}       
\usepackage{microtype}      
\usepackage{color}
\usepackage{xcolor}
\graphicspath{ {./images/} }
\usepackage{multirow}

\usepackage{sidecap}

\makeatletter
\renewcommand*{\@fnsymbol}[1]{\ensuremath{\ifcase#1\or *\or \dagger\or \ddagger\or
    \mathsection\or \mathparagraph\or \|\or **\or \dagger\dagger
    \or \ddagger\ddagger \else\@ctrerr\fi}}
\makeatother

\begin{document}

\pagestyle{headings}
\mainmatter
\def\ECCVSubNumber{636}  

\title{DiVA: Diverse Visual Feature Aggregation for Deep Metric Learning} 

\titlerunning{DiVA: Diverse Visual Feature Aggregation for Deep Metric Learning}
%
\author{Timo Milbich\thanks{Equal first and last authorship.}\inst{1} \and
Karsten Roth$^*$\inst{1,2} \and
Homanga Bharadhwaj\inst{3,4} \and
Samarth Sinha\inst{2,4} \and
Yoshua Bengio\inst{2,5} \and
Björn Ommer$^*$\inst{1} \and
Joseph Paul Cohen$^*$\inst{2}
}
\authorrunning{T. Milbich, K. Roth et al.}
%
\institute{Heidelberg Collaboratory for Image Processing (HCI), Heidelberg University  \and
Mila, Universite de Montreal \and Vector Institute, Toronto Robotics Institute \and
University of Toronto \and CIFAR Senior Fellow}
\maketitle

\setcounter{footnote}{0}

\begin{abstract}
Visual similarity plays an important role in many computer vision applications. Deep metric learning (DML) is a powerful framework for learning such similarities which not only generalize from training data to identically distributed test distributions, but in particular also translate to \textit{unknown} test classes. However, its prevailing learning paradigm is class-discriminative supervised training, which typically results in representations specialized in separating training classes. For effective generalization, however, such an image representation needs to capture a diverse range of data characteristics. To this end, we propose and study multiple complementary learning tasks, targeting conceptually different data relationships by only resorting to the available training samples and labels of a standard DML setting. Through simultaneous optimization of our tasks we learn a single model to aggregate their training signals, resulting in strong generalization and state-of-the-art performance on multiple established DML benchmark datasets. Code can be found here: \url{https://github.com/Confusezius/ECCV2020_DiVA_MultiFeature_DML}
\end{abstract}

\section{Introduction}
Many applications in computer vision, such as image retrieval~\cite{proxynca,dvml,margin} and face verification~\cite{semihard,npairs}, rely on capturing visual similarity, where approaches are commonly driven by Deep Metric learning (DML)~\cite{npairs,margin,proxynca}. These models aim to learn an embedding space which meaningfully reflects similarity between training images and, more importantly, generalizes to test classes which are \textit{unknown} during training. Even though models are evaluated on transfer learning, the prevailing training paradigm in DML utilizes discriminative supervised learning. Consequently, the learned embedding space is specialized to features which help only in separating among training classes, and may not correctly translate to unseen test classes. 
Now, if supervised learning does not result in sufficient generalization, how can we exploit the available training data and class labels to provide additional training signals beyond the standard discriminative task? 
\\
Recent breakthroughs in self-supervised learning have shown that contrastive image relations inferred from images themselves yield rich feature representations which even surpass the ability of supervised features to generalize to novel downstream task~\cite{cpc,moco,newhintonpaper}. However, although DML typically also learns from image relations in the form of pairs~\cite{contrastive}, triplets~\cite{margin,semihard} or more general image tuples~\cite{lifted,quadtruplet}, the complementary benefit of self-supervision in DML is largely unstudied. Moreover, the commonly available class assignments give rise to image relations aside from the standard, supervised learning task of `pulling' samples with identical class labels together while `pushing' away samples with different labels. As such ranking-based learning is not limited to discriminative training only, other relations can be exploited to learn beneficial data characteristics which so far have seen little coverage in DML literature.
\\
In this paper, we tackle the issue of generalization in DML by designing diverse learning tasks complementing standard supervised training, leveraging only the commonly provided training samples and labels. Each of these tasks aims at learning features representing different relationships between our training classes and samples: \textit{(i)} features discriminating among classes, \textit{(ii)} features shared across different classes, \textit{(iii)} features capturing variations within classes and \textit{(iv)} features contrasting between individual images. Finally, we present how to effectively incorporate them in a unified learning framework. In our experiments we study mutual benefits of these tasks and show that joint optimization of diverse representations greatly improves generalization performance as shown in Fig.~\ref{fig:first_page} (left), outperforming the state-of-the-art in DML. Our contributions can be summarized as follows:

\begin{figure}[t]
    \centering
    \includegraphics[width=0.8\linewidth]{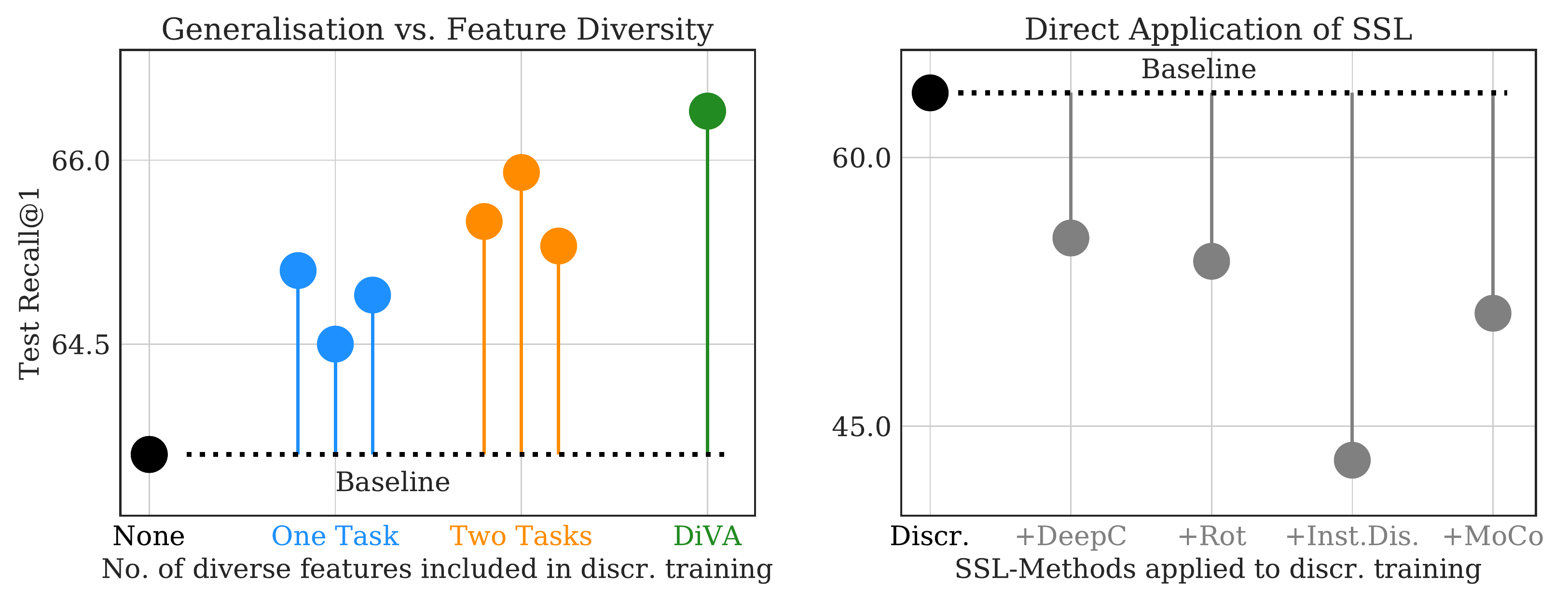}
    \caption{\textit{DML using diverse learning tasks vs. direct incorporation of self-supervision.} (Left) Generalization performance increases with each task added to training, independent of the exact combination of our proposed tasks (blue: one extra task, orange: two extra tasks, green: all tasks). (Right) Directly combining supervised learning with self-supervised learning techniques such as DeepC(luster)~\cite{deepcluster}, Rot(Net)~\cite{predict_rotations}, Inst.Dis(crimination)~\cite{wu2018unsupervised} or Mo(mentum)Co(ntrast)~\cite{moco} actually hurts DML generalization.}
    %
    \label{fig:first_page}
\end{figure}

\begin{itemize}
    \item We design novel triplet learning tasks resulting in a diverse set of features and study their complementary impact on supervised DML.
    \item We adopt recent contrastive self-supervised learning to the problem of DML and extend it to effectively support supervised learning, as direct incorporation of self-supervised learning does not benefit DML (cf. Fig\ref{fig:first_page}) (right).
    \item We show how to effectively incorporate these learning tasks in a single model, resulting in state-of-the-art performance on standard DML benchmark sets.
\end{itemize}

\section{Related Work}
\label{sec:Related_Works} 
\noindent
\textbf{Deep Metric Learning.} 
Deep Metric Learning is one of the primary frameworks for image retrieval \cite{proxynca,dvml,mic,margin}, zero-shot generalization \cite{semihard,mic,Sanakoyeu_2019_CVPR,cliquecnn} or face verification \cite{face_verfication_inthewild,sphereface,arcface}. It is also closely related to recent successful unsupervised representation learning approaches employing contrastive learning \cite{moco,pretextmisra,newhintonpaper}. Commonly, DML approaches are formulated as ranking tasks on data tuples such as image pairs\cite{contrastive}, triplets\cite{face_verfication_inthewild,margin}, quadruplets\cite{quadtruplet} or higher order relations\cite{npairs,lifted,multisimilarity}. Effective training of these methods is typically promoted by tuple mining strategies alleviating the high sampling complexity, such as distance-based \cite{semihard,margin}, hierarchical~\cite{htl} or learned~\cite{Roth_2020_CVPR}. Methods like ProxyNCA\cite{proxynca}, Softtriple\cite{arcface}, Arcface\cite{arcface} or Normalized Softmax\cite{zhai2018classification} introduce learnable data proxies which represent entire subsets of the data, thus circumventing the tuple selection process. Orthogonally, DML research has started to pay more emphasis on the training process itself. This involves the generation of artificial training data \cite{dvml,hardness-aware} or adversarial objectives \cite{daml}. MIC~\cite{mic} explains away intra-class variance to strengthen the discriminative embedding. \cite{Sanakoyeu_2019_CVPR} propose to separate the input data space to learn subset-specific, yet still only class-discriminative representations similar to other ensemble methods~\cite{dreml,abe,abier,milbich_pami_20}. In contrast, we learn different embeddings on conceptually different tasks to capture diverse image features.
\\
\noindent
\textbf{Self-supervised Representation Learning.}
Commonly, self-supervised representation learning aims to learn transferable feature representations from unlabelled data, and is typically applied as pre-training for downstream tasks\cite{hsu2018unsupervised,pr20_reliable_relations}. Early methods on representation learning are based on sample reconstructions \cite{vincentnoisyauto,vae} which have been further extended by interpolation constraints \cite{acai} and generative adversarial networks \cite{chen2016infogan,ali,hali,bigan}. Further, introducing manually designed surrogate objectives encourage self-supervised models to learn about data-related properties. Such tasks range from predicting image rotations \cite{predict_rotations}, solving a Jigsaw puzzle \cite{jigsaw,jigsaw++,buechler_ECCV_2018} to iteratively refining the initial network bias using clustering algorithms \cite{deepcluster}. Recently, self-supervision approaches based on contrastive learning result in strong features performing close to or even stronger than supervised pretraining \cite{moco,pretextmisra,newhintonpaper,tian2019contrastive} by leveraging invariance to realistic image augmentations. As the these approaches are essentially defined on pairwise image relations, they share common ground with ranking-based DML. In our work, we extend such a contrastive objective to effectively complement supervised DML training. 
\\
\noindent
\textbf{Multi-task Learning.}
Concurrently solving different tasks is also employed by classical multi-task learning which are often based on a divide-and-conquer principle with multiple learner optimizing a given subtask. \cite{mtl_bhattarai} utilizes additional training data and annotations to capture extra information, while our tasks are defined on standard training data only. \cite{mtl_pu} learn different classifiers for groups of entire categories, thus following a similar motivation as some DML approaches~\cite{abier,Sanakoyeu_2019_CVPR}. The latter aims at learning more fine-grained, yet only discriminative features by reducing the data variance for each learner, thus being related to standard hard-negative mining~\cite{semihard}. In contrast, our work formulates various specific learning tasks to target \textit{different} data characteristics of the training data.


\section{Method}
\label{sec:Method}

Let $f_i := f(I_i, \theta) \in \mathbb{R}^N$ be a $N$-dimensional encoding of an image $I_i \in \mathcal{I}$ represented by a deep neural network with parameters $\theta$. Based on $f_i$, deep metric learning (DML) aims to learn image embeddings $\phi_i := \phi(f_i,\zeta): \mathbb{R}^N \mapsto \mathbb{R}^D$ which allow to measure the similarity between images $I_i,I_j$ as $d^\phi_{i,j} := d(\phi_i,\phi_j)$ under a predefined distance metric $d(\cdot,\cdot)$. Typically, $\phi(\cdot,\zeta)$ is a linear layer on the features $f$ representation, parameterized by $\zeta$ and normalized to the real hypersphere $\mathbb{S}^{D-1}$ for regularization~\cite{margin}. $d(\cdot,\cdot)$ is usually chosen to be the Euclidean distance. In standard supervised DML, $\phi$ is then optimized to reflect semantic similarity between images $I_i$ defined by the corresponding class labels $y_i \in \mathcal{Y}$.
\\
While there are many ways to define training objectives on $\phi$, ranking losses, such as variants of the popular triplet loss~\cite{margin,npairs,lifted}, are a natural surrogate for the DML problem. Based on image triplets $t = \{I_a, I_p, I_n\}$ with $I_a$ defined as anchor, $I_p$ as a similar, positive and $I_n$ as a negative image, we minimize 
\begin{equation}
    \mathcal{L}_{\text{tri}}(t) = [d^\phi_{a,p} - d^\phi_{a,n} + \gamma]_+ \;,
\label{eq:triplet}
\end{equation}
\noindent
where $[\cdot]_+$ defines the hinge function which clips any negative value to zero. Hence, we maximize the gap between $d^\phi_{a,p}$ and $d^\phi_{a,n}$ as long as a margin $\gamma$ is violated. 
\\
In supervised DML, $\mathcal{L}_{\text{tri}}(t)$ is typically optimized to discriminate between classes. Thus, $f$ is trained to predominantly capture highly discriminative features while being invariant to image characteristics which do not facilitate training class separation. However, as we are interested in generalizing to unknown test distributions, we should rather aim at maximizing the amount of features captured from the training set $\mathcal{I}$, i.e. additionally formulate learning tasks which yield features beyond mere class-discrimination.
\\
In order to formulate such tasks, we make use of the fact that triplet losses are instance-based. Thus, conceptually they are not restricted to class discrimination, but allow to learn \textit{commonalities between the provided anchor $I_a$ and positive $I_p$} compared to a negative sample $I_n$. Following, we will use this observation to also learn those commonalities which are neglected by discriminative training, such as commonalities shared between and within classes.

\subsection{Diverse learning tasks for DML}
\label{sec:feature_tasks}
We now introduce several tasks for learning a diverse set of features, resorting only to the standard training information provided in a DML problem. Each of these tasks is designed to learn features which are conceptually neglected by the others to be mutually complementary. 
First, we introduce the intuition behind each feature type, before describing how to learn it based on pairwise or triplet-based image relations. 
\\
\begin{figure}[t]
    \centering
    \includegraphics[width=0.85\linewidth]{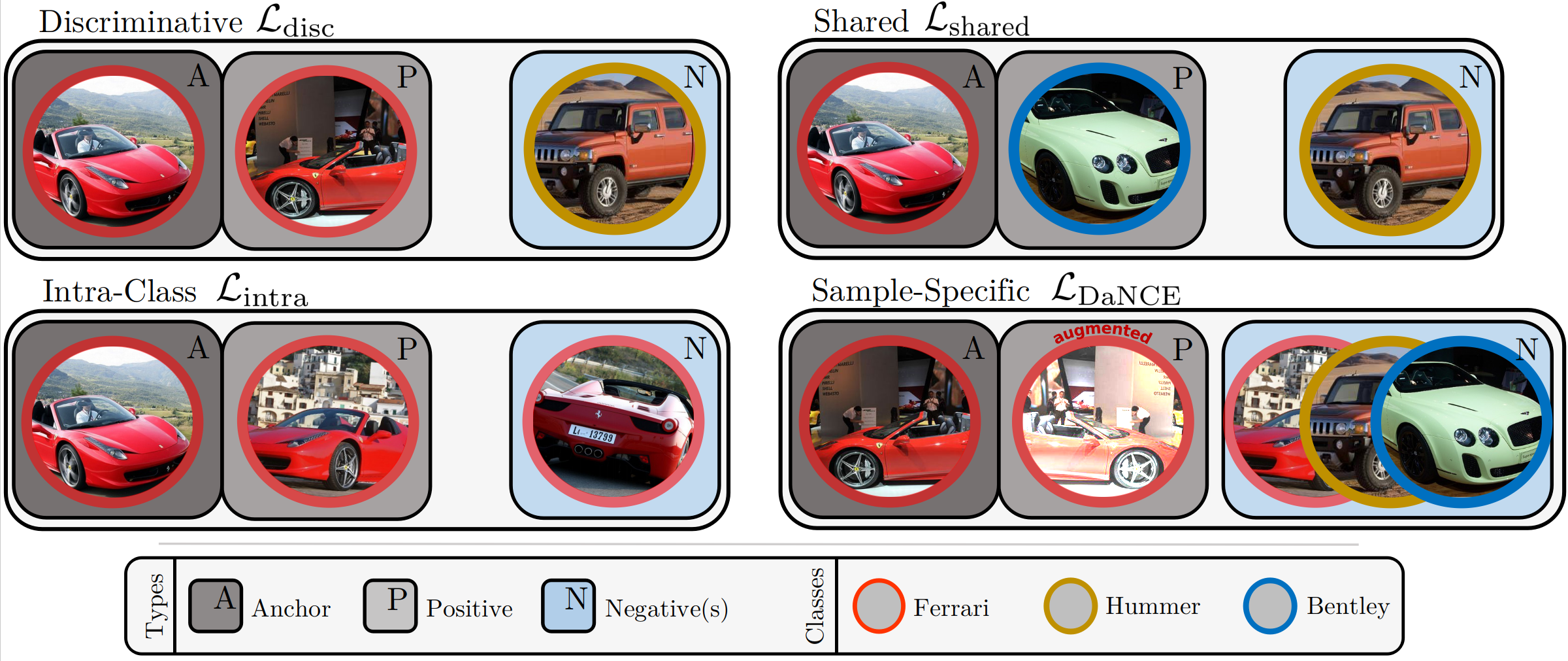}
    \caption{\textit{Schematic description of each task.} We learn four complementary tasks to capture features focusing on different data characteristics. The standard \textit{class-discriminative} task which learning features separating between samples of different classes, the \textit{shared} task which captures features relating samples across different classes, a \textit{sample-specific} task to enforce image representations invariant to transformations and finally the \textit{intra-class} task modelling data variations within classes.}
    \label{fig:tasks}
\end{figure}

\noindent
\textbf{Class-discriminative features.} These features are learned by standard class-discriminative optimization of $\phi$ and focus on data characteristics which allow to accurately separate one class from all others. It is the prevailing training signal of common classification-based~\cite{angular,arcface,zhai2018classification}, proxy-based~\cite{proxynca,softriple} or ranking-based~\cite{mic,abier} approaches. For the latter, we can formulate the training task using Eq.~\ref{eq:triplet} by means of triplets $\{I_a,I_p,I_n\} \in \mathcal{T}_{\text{disc}}$ with $y_a=y_p$ and $y_a\neq y_n$, as 

\begin{equation}
    \mathcal{L}_{\text{disc}} = \frac{1}{Z}\sum_{t \sim \mathcal{T}_{\text{disc}}}\mathcal{L}_{\text{tri}}(t) \;,
\end{equation}

\noindent
thus minimizing embedding distances between samples of the same class while maximizing it for samples of different classes. Moreover, the discriminative signal is important to learn how to aggregate features into classes, following the intuition of ``the whole is more than the sum of its parts" analyzed in Gestalt theory~\cite{gestalt}.
\\
\textbf{Class-shared features.} 
In contrast to discriminative features which look for characteristics \textit{separating classes}, class-shared features capture commonalities, i.e variations, \textit{shared across classes}. For instance, cars have a certain invariance towards color changes, thus being of little help when separating between them. However, to learn about this characteristic is actually beneficial when generalizing to other colorful object classes like flowers or fishes. Given suitable label information, learning such features would naturally follow the standard discriminative training setup. 
However, having only class labels available, we must resort to approximations. To this end, we exploit the hypothesis that for most arbitrarily sampled triplets $\{I_a, I_p, I_n\} \in \mathcal{T}_{\text{shared}}$ with each constituent coming from mutually different classes, i.e. $y_a \neq y_p \neq y_n$, the anchor $I_a$ and positive $I_p$ share some common pattern when compared the negative image $I_n$. Commonalities which are frequently observed between classes $y_a, y_p$, will occur more often than noisy patterns which are unique to few $t_{\text{shared}}$, which is commonly observed when learning on imbalanced data \cite{imb1,imb2,imb3}. Learning is then performed by optimizing

\begin{equation}
    \mathcal{L}_{\text{shared}} = \frac{1}{Z}\sum_{t \sim \mathcal{T}_{\text{shared}}}\mathcal{L}_{\text{tri}}(t) \;.
\end{equation}

\noindent
As deep networks learn from frequent backpropagation of similar learning signals resulting in informative gradients, only prominent shared features are captured. Further, since shared features can be learned between any classes, we need to warrant diverse combinations of classes in our triplets $\mathcal{T}_{\text{shared}}$. Thus, enabling triplet constituents to be sampled from the whole embedding space $\phi$ using distance-based sampling~\cite{margin} is crucial to avoid any bias towards samples which are mostly far (random sampling) or close (hard-negative sampling) to a given anchor $I_a$. 
\\
\textbf{Intra-class features.}
The tasks defined so far model image relations across classes. In contrast, intra-class features describe variations within a given class. While these variations may also apply to other classes (thus exhibiting a certain overlap with class-shared features), more class-specific details are targeted. Hence, to capture such data characteristics by means of triplet constraints, we can follow a similar intuition as for learning class-shared features. Thus, to learn intra-class features, we define triplets by means of triplets $\{I_a, I_p, I_n\} \in \mathcal{T}_{\text{intra}}$ with $y_a = y_p = y_n$, i.e. each constituent coming from the same class, and minimize 

\begin{equation}
    \mathcal{L}_{\text{intra}} = \frac{1}{Z}\sum_{t \sim \mathcal{T}_{\text{intra}}}\mathcal{L}_{\text{tri}}(t) \;.
\end{equation}

\noindent
\textbf{Sample-specific features.} 
Recent approaches for self-supervised learning~\cite{cpc,moco,bachman2019amdim} based on noise contrastive estimation (NCE)~\cite{nce} show that features exhibiting strong generalization for transfer learning can be learned only from training images themselves. As NCE learns to increase the correlation between embeddings of an anchor sample and a similar positive sample by constrasting against a set of negative samples, it naturally translates to DML. He et al.~\cite{moco} proposed an efficient self-supervised framework which first applies data augmentation to generate positive surrogates $\tilde{I}_a$ for a given anchor $I_a$. Next, using NCE we contrast their embeddings $\phi_a:=\phi(I_a,\zeta), \tilde{\phi}_a:=\phi(\tilde{I_a},\zeta)$ against randomly sampled negatives $I_n \in \mathcal{N} \subset \mathcal{I}$ by minimizing

\begin{equation}
    \mathcal{L}_{\text{NCE}} = \frac{1}{Z}\sum_{I_a \sim \mathcal{I}} -\log\frac{\exp(\phi(I_a,\zeta)^\top \phi(\tilde{I}_a,\zeta)/\tau)}{\sum_{I_n \in \mathcal{N}} \exp(\phi(I_a,\zeta)^\top \phi(I_n,\zeta) / \tau)} 
    \label{eq:nce}
\end{equation}

\noindent
where the temperature parameter $\tau$ is adjusted during optimization to control the training signal, especially during earlier stages of training. By contrasting each sample against many negatives, i.e. large sets $\mathcal{N}$, this task effectively yields a general, class-agnostic features description of our data. Moreover, as the contrastive objective explicitly increases the similarity of an anchor image with its augmentations, invariance against data transformations and scaling are learned. Fig. \ref{fig:tasks} summarizes and visually explains the different training objectives of each task.

\subsection{Improved generalization by multi feature learning}
\label{sec:Method_Details}

Following we show how to efficiently incorporate the learning tasks introduced in the previous section into a single DML model. We first extend the objective Eq.~\ref{eq:nce} using established triplet sampling strategies for improved adjustment to DML, before we jointly train our learning tasks for maximal feature diversity.  
\\
\\
\noindent
\textbf{Adapting noise contrastive estimation to DML.}
Efficient strategies for mining informative negatives $I_n$ are a key factor~\cite{semihard} for successful training of ranking-based DML models. Since NCE essentially translates to a ranking between images $I_a,\tilde{I}_a,I_n$, its learning signal is also impaired if $I_n \in \mathcal{N}$ are uninformative, i.e. $d^\phi(I_a,I_n)$ being large. To this end, we control the contribution of each negative $I_n \in \mathcal{N}$ to $\mathcal{L}_{\text{NCE}}$ by a weight factor $w(d) = \min(\lambda, q^{-1}(d))$. Here, $q(d) = d^{D-2}\left[1-\frac{1}{4}d^2\right]^{\frac{D-3}{2}}$ is the distribution of pairwise distances\footnote{To compute $d$, we use the euclidean distance between samples. Since $\phi$ is regularized to the unit hypersphere $\mathbb{S}^{D-1}$, the euclidean distance correlates with cosine distance.} on the $D$-dimensional unit hypersphere $\mathbb{S}^{D-1}$ and $\lambda$ a cut-off parameter. Similar to \cite{margin}, $w(d)$ helps to equally weigh negatives from the whole range of possible distances in $\phi$ and, in particular, increases the impact of harder negatives. Thus, our distance-adapted NCE loss becomes

\begin{equation}
    \mathcal{L}_{\text{DaNCE}} = \frac{1}{Z}\sum_{I_a \sim \mathcal{I}} -\log\frac{\exp(\phi(I_a)^\top \phi(\tilde{I}_a)/\tau)}{\sum_{I_n \in \mathcal{N}} \exp(w(d^\phi_{a,n}) \cdot \phi^*(I_n)^\top \phi(I_a) / \tau)} \; .
    \label{eq:dance}
\end{equation}

\noindent
NCE-based objectives learn best using large sets of negatives~\cite{moco}. However, naively utilizing only negatives from the current mini-batch constrains $\mathcal{N}$ to the available GPU memory. To alleviate this limitation, we follow \cite{moco} and realize $\mathcal{N}$ as a large memory queue, which is constantly updated with embeddings $\phi^*(\tilde{I}_a)$ from training iteration $t$ by utilising the running-average network $\phi^{*,{t+1}} = \mu\phi^{*,t} + (1-\mu)\phi^{t}$.
\begin{figure}[t]
    \centering
    \includegraphics[width=0.85\linewidth]{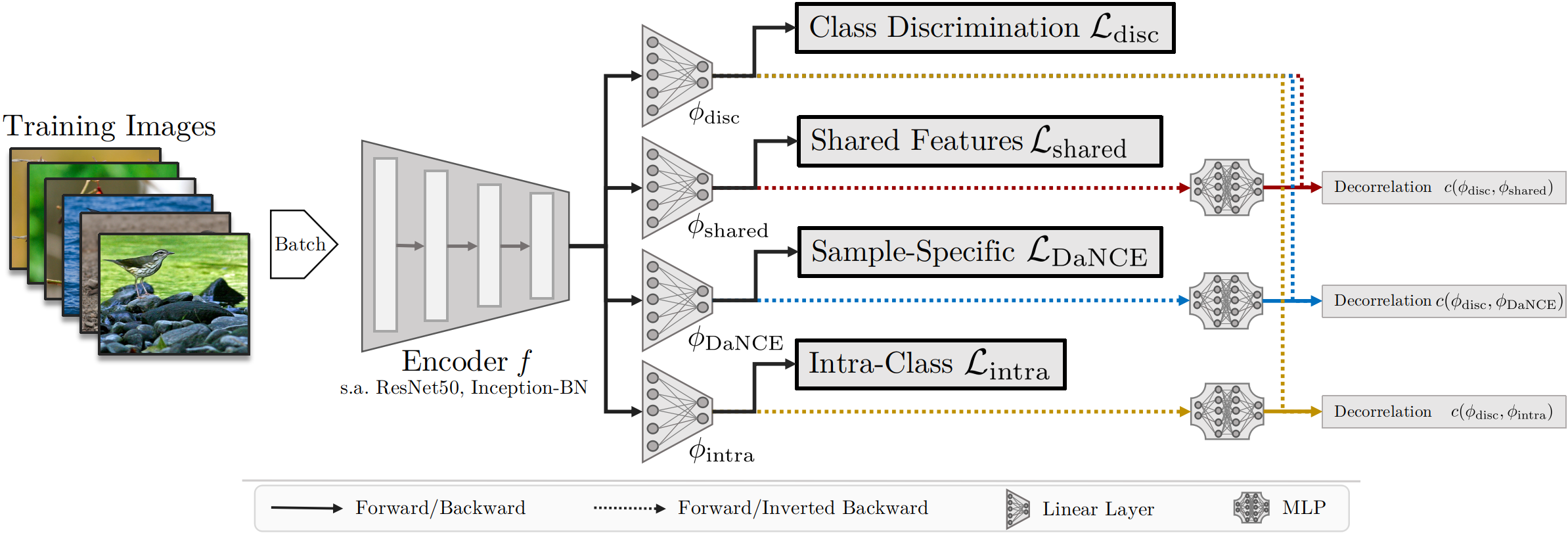}
    \caption{\textit{Architecture of our propose model.} Each task $\mathcal{L}_{\bullet}$ optimizes an individual embedding $\phi_{\bullet}$ implemented as a linear layer with a shared underlying feature encoder $f$. Pairwise decorrelation $c(\cdot,\cdot)$ of the embeddings utilizing the mapping $\psi$ based on a two-layer MLP encourages each task to further emphasize on its targeted data characteristics. Gradient inversion $R$ is applied during the backward pass to each embedding head. 
    %
}
    \label{fig:setup}
\end{figure}
\\
\\
\textbf{Joint optimization for maximal feature diversity.}
The tasks presented in Sec.~\ref{sec:feature_tasks} are formulated to extract mutually complementary information from our training data. In order to capture their learned features in a single model to obtain a rich image representation, we now discuss how to jointly optimize these tasks. 
\\
While each task targets a semantically different concept of features, their driving learning signals are based on potentially contradicting ranking constraints on the learned embedding space. Thus, aggregating these signals to optimizing a joint, single embedding function $\phi$ may entail detrimental interference between them. In order to circumvent this issue, we learn a dedicated embedding space for each task, as often conducted in multi-task optimization~\cite{mic,fasterrcnn}, i.e.  $\phi_{\text{disc}}(f),\phi_{\text{shared}}(f),\phi_{\text{intra}}(f)$ and $\phi_{\text{nce}}(f)$ with $\phi_{\bullet}(f): \mathbb{R}^N \mapsto \mathbb{R}^D$ (cf. Sec.~\ref{sec:Method}).
As all embeddings share the same feature extractor $f$, each task still benefits from the aggregated learning signals. Additionally, as there may still be redundant overlap in the information captured by each task, we mutually decorrelate these representations, thus maximizing the diversity of the overall training signal. Similar to \cite{abier,mic} we minimize the mutual information of two embedding functions $\phi^a$, $\phi^b$ by maximizing their correlation $c$ in the embedding space of $\phi^b$, followed by a gradient reversal. For that, we learn a mapping $\psi: \mathbb{R}^D \mapsto \mathbb{R}^D$ from $\phi_i^a$ to $\phi_i^b$ given an image $I_i$ and compute the correlation $c(\phi_i^a, \phi_i^b) = \lVert (R(\phi_{i}^a) \odot \psi(R(\phi_i^b))) \rVert_2^2$ with $\odot$ being the point-wise product. $R$ denotes a gradient reversal operation, which inverts the resulting gradients during backpropagation. Maximizing $c$ results in $\psi$ aiming to make $\phi^a$ and $\phi^b$ comparable. However, through the subsequent gradients reversal, we actually decorrelate the embedding functions. Joint training of all tasks is finally performed by minimizing
\begin{align}
\mathcal{L} &= \mathcal{L}_{\text{disc}} + \alpha_1\mathcal{L}_{\text{shared}} + \alpha_2\mathcal{L}_{\text{intra}} + \alpha_3\mathcal{L}_{\text{DaNCE}} -\sum_{(\phi_a,\phi_b) \in \mathcal{P}} \rho_{a,b} \cdot c(\phi_a,\phi_b) 
\label{eq:joint}
\end{align}
\noindent
where $\mathcal{P}$ denotes the pairs of embeddings to be decorrelated. We found 
\begin{equation}\label{eq:pairs}
\mathcal{P} = \{(\phi_\text{disc},\phi_\text{DaNCE}),(\phi_\text{disc},\phi_\text{shared}),(\phi_\text{disc},\phi_\text{intra})\} \; ,
\end{equation}
to work best, which decorrelates the auxiliary tasks with the class-discriminative task. Initial experiments showed that further decorrelation $c(\bullet,\bullet)$ among the auxiliary tasks does not result in further benefit and is therefore disregarded. The weighting parameters $\rho_{a,b}$ adjusting the degree of decorrelation between the embeddings are set to the same, constant value in our implementation. Fig.~\ref{fig:setup} provides an overview of our model. Finally, we combine our learned embedding to form an ensemble representation to fully make use of all information.
\\
\noindent
\textbf{Computational costs.}
We train all tasks using the same mini-batch to avoid computational overhead. While optimizing each learner on an individual batch can further alleviate training signal interference~\cite{Sanakoyeu_2019_CVPR,dreml}, training time increases significantly. Using a single batch per iteration, we minimize the required extra computations to the extra forward pass through $\phi^*$ (however without computing gradients) for contrasting against negatives sampled from the memory queue as well as the small mapping networks $\psi$. Across datasets, we measure an increase in training time by $10-15\%$ per epoch compared to training a standard supervised DML task. This is comparable to or lower than other methods, which perform a full clustering on the dataset~\cite{mic,Sanakoyeu_2019_CVPR} after each epoch, compute extensive embedding statistics~\cite{horde} or simultaneously train generative models~\cite{dvml}.


\section{Experiments}
\label{sec:Experiments}
Following we first present our implementation details and the benchmark datasets. Next, we evaluate our proposed model and study how our learning tasks complement each other and improve over baseline performances. Finally, we discuss our results in the context of the current state-of-the-art and conduct analysis and ablation experiments.
\\
\textbf{Implementation details.} 
We follow the common training protocol of \cite{margin,mic,Sanakoyeu_2019_CVPR} for implementations utilizing a ResNet50-backbone. The shorter image axis is resized to $256$, followed by a random crop to $224\times 224$ and a random horizontal flip with $p=0.5$. During evaluation, only a center crop is taken after resizing. The embedding dimension is set to $D=128$ for each task embedding. For model variants using the Inception-V1 with Batch-Normalization\cite{googlenetv2}, we follow \cite{multisimilarity,horde} and use $D=512$. Resizing, cropping and flipping is done in the same way as for ResNet50 versions. The implementation is done using the PyTorch framework\cite{pytorch}, and experiments are performed on compute clusters containing NVIDIA Titan X, Tesla V4, P100 and V100, always limited to 12GB VRAM following the standard training protocol~\cite{margin}. For DiVA, we utilise the triplet-based margin loss~\cite{margin} with fixed margin $\gamma=0.2$ and $\beta=1.2$ and fixed temperature $\tau=0.1$. 
\\
\textbf{Hyperparameters.}
For training, we use Adam\cite{adam} with learning rate $10^{-5}$ and a weight decay of $5\cdot 10^{-4}$. For ablations, we use no learning rate scheduling, while our final model is trained using scheduling values determined by cross-validation. We train for $150$ epochs. Our joint training framework can be adjusted by setting the de-correlation weights $\rho_{a,b}$ and weight parameters $\alpha_i$. In both cases we utilize the same values for all learning task, thus we effectively only adjust two parameters $[\rho,\alpha]$ to each benchmark sets: CUB200-2011 (IBN: $[300,0.15]$, R50 $[1500,0.3]$), CARS196 (IBN: $[100,0.15]$, R50 $[100,0.1]$), SOP (IBN: $[150,0.2]$, R50 $[150,0.2]$). This is comparable to other approaches, e.g. MS~\cite{multisimilarity}, SoftTriple~\cite{softriple}, D\&C~\cite{Sanakoyeu_2019_CVPR}. As our auxiliary embeddings generalize better, they are more emphasized for computing $d_{i,j}$ during testing (e.g. double on CUB200 and CARS196).
\\
\textbf{Datasets.} We evaluate the performance on three common benchmark datasets with standard training/test splits (see e.g. \cite{margin,mic,Sanakoyeu_2019_CVPR,multisimilarity}):
\textit{CARS196}\cite{cars196}, which contains 16,185 images from 196 car classes. The first 98 classes containing 8054 images are used for training, while the remaining 98 classes with 8131 images are used for testing. \textit{CUB200-2011}\cite{cub200-2011} with 11,788 bird images from 200 classes. Training/test sets contain the first/last 100 classes with 5864/5924 images respectively. \textit{Stanford Online Products (SOP)}\cite{lifted} provides 120,053 images divided in 22,634 product classes. 11318 classes with 59551 images are used for training, while the remaining 11316 classes with 60502 images are used for testing.

 \begin{table*}[t]
   \label{tab:baselines}
   \setlength\tabcolsep{1.4pt}
   \centering
   \begin{tabular}{l|c|ccc|ccc|ccc}
     \toprule
     \multicolumn{2}{l}{Dataset $\rightarrow$} & \multicolumn{3}{c}{CUB200-2011\cite{cub200-2011}} & \multicolumn{3}{c}{CARS196\cite{cars196}} & 
     \multicolumn{3}{c}{SOP\cite{lifted}} \\
     \midrule
     Approach $\downarrow$ & Dim & R@1 & R@2 & NMI & R@1 & R@2 & NMI & R@1 & R@10 & NMI\\
     \midrule
     Margin\cite{margin} (orig, R50)  & 128 & 63.6 & 74.4 & 69.0 & 79.6 & 86.5 & 69.1 & 72.7 & 86.2 & 90.7 \\
     Margin\cite{margin} (ours, IBN) & 512 & 63.6 & 74.7 & 68.3 & 79.4 & 86.6 & 66.2 & 76.6 & 89.2 & 89.8 \\
     \hline
     DiVA (IBN, D \& Da)      & 512 & 64.5 & 76.0 & 68.8 & 80.4 & 87.7 & 67.2 & 77.0 & 89.4 & \textbf{90.1}\\
     DiVA (IBN, D \& S)       & 512 & 65.1 & 76.4 & 69.0 & 81.5 & 88.3 & 66.8 & 77.2 & 89.6 & 90.0\\ 
     DiVA (IBN, D \& I)       & 512 & 64.9 & 75.8 & 68.4 & 80.6 & 87.9 & 67.4 & 76.9 & 89.4 & 89.9\\
     DiVA (IBN, D \& Da \& I) & 510 & 65.3 & 76.5 & 68.3 & 82.2 & 89.1 & 67.8 & 75.8 & 89.0 & 89.8 \\ 
     DiVA (IBN, D \& S \& I)  & 510 & 65.5 & 76.4 & 68.4 & 82.1 & 89.4 & 67.2 & 77.0 & 89.3 & 89.7 \\      
     DiVA (IBN, D \& Da \& S) & 510 & 65.9 & 76.7 & 68.9 & 82.6 & 89.6 & 68.0 & 77.4 & 89.6 & 90.1 \\  
     \hline
     DiVA (IBN, D \& Da \& S \& I) & 512 & \textbf{66.4} & \textbf{77.2} & \textbf{69.6} & \textbf{83.1} & \textbf{90.0} & \textbf{68.1} & \textbf{77.5} & \textbf{90.3} & \textbf{90.1} \\    
     \bottomrule
   \end{tabular}
     \caption{\textit{Comparison of different combinations of learning tasks.} I(nception-V1) B(atch-)N(ormalization), and R(esNet)50 denote the backbone architecture. No learning rate scheduling is used. Our tasks are denoted by D(\textit{iscriminative}), S(\textit{hared}), I(\textit{ntra-Class}) \& and Da(\textit{NCE}). The dimensionality per task embedding depends on the number of tasks used, totalling in $D=512$ (two tasks use $256$ each, three $170$, four $128$.}
     \label{tab:baselines}
 \end{table*}

\subsection{Performance study of multi-feature DML}\label{sec:perf_study}
We now compare our model and the complementary benefit of our proposed feature learning tasks for supervised DML. Tab.~\ref{tab:baselines} evaluates the performance of our model based on margin loss~\cite{margin}, a triplet based objective with an additionally learnable margin, and distance-weighted triplet sampling \cite{margin}. We use Inception-V1 with Batchnorm and a maximal aggregated embedding dimensionality of $512$. Thus, if two tasks are utilized, each embedding has $D=256$, in case of three tasks $170$ and four tasks result in $D=128$. No learning rate scheduling is used. Evaluation is conducted on CUB200-2011~\cite{cub200-2011}, CARS196~\cite{cars196} and SOP~\cite{lifted}. Retrieval performance is measured through Recall@k\cite{recall} and clustering quality via Normalized Mutual Information (NMI)~\cite{nmi}. While our results vary between possible task combinations, we observe that the generalization of our model consistently increases with each task added to the joint optimization. Our strongest model including all proposed tasks improves the generalization performance by 2.8\% on CUB200-2011, 3.7\% on CARS196 and 0.9\% on SOP. This highlights that \textit{(i)} purely discriminative supervised learning disregards valuable training information and \textit{(ii)} carefully designed learning tasks are able to capture this information for improved generalization to unknown test classes. We further analyze our observations in the ablation experiments.

\begin{table*}[t]
    \setlength\tabcolsep{1.5pt}
    \centering
     \begin{tabular}{l|c|cc|c||cc|c||cc|c}
        \toprule
        Dataset $\rightarrow$ & & \multicolumn{3}{c}{CUB200-2011\cite{cub200-2011}} & \multicolumn{3}{c}{CARS196\cite{cars196}}
        & \multicolumn{3}{c}{SOP\cite{lifted}}\\ 
        \hline
        Approach $\downarrow$ & Dim & R@1 & R@2 & NMI & R@1 & R@2 & NMI & R@1 & R@2 & NMI \\
        \midrule
        HTG\cite{htg}             & 512 & 59.5 & 71.8 & - & 76.5 & 84.7 & - & - & - & -\\
        HDML\cite{hardness-aware} & 512 & 53.7 & 65.7 & 62.6 & 79.1 & 87.1 & 69.7 & 68.7 & 83.2 & 89.3\\
        Margin\cite{margin}       & 128 & 63.6 & 74.4 & 69.0 & 79.6 & 86.5 & 69.1 & 72.7 & 86.2 & 90.8\\
        HTL\cite{htl}             & 512 & 57.1 & 68.8 & - & 81.4 & 88.0 & - & 74.8 & 88.3 & -\\
        DVML\cite{dvml}           & 512 & 52.7 & 65.1 & 61.4 & 82.0 & 88.4 & 67.6 & 70.2 & 85.2 & 90.8\\
        MultiSim\cite{multisimilarity}  & 512 & 65.7 & 77.0 & -    & 84.1 & 90.4 & -    & 78.2 & 90.5 & -   \\
        D\&C\cite{Sanakoyeu_2019_CVPR}  & 128 & 65.9 & 76.6 & 69.6 & 84.6 & 90.7 & 70.3 & 75.9 & 88.4 & 90.2\\
        MIC\cite{mic}             & 128 & 66.1 & 76.8 & 69.7 & 82.6 & 89.1 & 68.4 & 77.2 & 89.4 & 90.0\\        
        \hline
        \multicolumn{11}{l}{Significant increase in network parameter:} \\      
        \hline
        HORDE\cite{horde}+Contr.\cite{contrastive} & 512 & 66.3 & 76.7 & - & 83.9 & 90.3 & -    & -    & -    & -   \\
        Softtriple\cite{softriple}                 & 512 & 65.4 & 76.4 & - & 84.5 & 90.7 & 70.1 & 78.3 & 90.3 & \textbf{92.0}\\
        \hline
        \multicolumn{11}{l}{Ensemble Methods:} \\
        \hline
        A-BIER\cite{abier}    & 512 & 57.5 & 68.7 & - & 82.0 & 89.0 & - & 74.2 & 86.9 & -\\        
        Rank\cite{rankedlist} & 1536 & 61.3 & 72.7 & 66.1 & 82.1 & 89.3 & 71.8 & \textbf{79.8} & \textbf{91.3} & 90.4\\
        DREML\cite{dreml}     & 9216 & 63.9 & 75.0 & 67.8 & 86.0 & 91.7 & \textbf{76.4} & -    & -    & -\\
        ABE\cite{abe}         & 512  & 60.6 & 71.5 & -    & 85.2 & 90.5 & -    & 76.3 & 88.4 & -\\
        \hline
        \multicolumn{11}{l}{Inception-BN} \\           
        \hline
        \textbf{Ours} (Trip-DiVA-IBN-512)   & 512  & 66.7 & 77.1 & 69.3 & 83.1 & 90.3 & 68.8 & 76.9 & 88.9 & 89.4\\            
        \textbf{Ours} (DiVA-IBN-512)    & 512  & 66.8 & 77.7 & 70.0 & 84.1 & 90.7 & 68.7 & 78.1 & 90.6 & 90.4\\ 
        \hline
        \multicolumn{11}{l}{ResNet50} \\           
        \hline
        \textbf{Ours} (Margin\cite{margin}-R50-512) & 512 & 64.4 & 75.4 & 68.4 & 82.2 & 89.0 & 68.1 & 78.3 & 90.0 & 90.1\\ 
        \textbf{Ours} (Trip-DiVA-R50-512) & 512 & 68.5 & 78.5 & \textbf{71.1} & \textbf{87.3} & \textbf{92.8} & 72.1 & 79.4 & 90.8 & 90.3\\         
        \textbf{Ours} (DiVA-R50-512) & 512 & \textbf{69.2} & \textbf{79.3} & \textbf{71.4} & \textbf{87.6} & \textbf{92.9} & 72.2 & \textbf{79.6} & \textbf{91.2} & 90.6\\ 
        \bottomrule
    \end{tabular}
    \caption{\textit{Comparison to the state-of-the-art methods on CUB200-2011}\cite{cub200-2011}, \textit{CARS196}\cite{cars196} \textit{and SOP}\cite{lifted}. DiVA-Arch-Dim describes the backbone used with DiVA (IBN: Inception-V1 with Batchnorm, R50: ResNet50) and the total training and testing embedding dimensionality. For fair comparison, we also ran a standard ResNet50 with embedding dimensionality of 512. Trip-DiVA-Arch-Dim indicates standard triplet loss as base objective. 
    }
    \vspace*{-1cm}
    \label{tab:sota}
\end{table*}

\subsection{Comparison to state-of-the-art approaches}
Next, we compare our model using fixed learning rate schedules per benchmark to the current state-of-the-art approaches in DML. For fair comparison to the different methods, we report result both using Inception-BN (IBN) and ResNet50 (R50) as backbone architecture. As Inception-BN is typically trained with embedding dimensionality of 512, we restrict each embedding to $D=128$ for direct comparison with non-ensemble methods. Thus we deliberately impair the potential of our model due to a significantly lower capacity per task, compared to the standard $D=512$. For comparison with ensemble approaches and maximal performance, we use a ResNet50~\cite{margin,mic,Sanakoyeu_2019_CVPR} architecture and the corresponding standard dimensionality $D=128$ per task. Fig.~\ref{tab:sota} summarizes our results using both standard triplet loss~\cite{semihard} and margin loss~\cite{margin} as our base training objective. In both cases, we significantly improve over methods with comparable backbone architectures and achieve new state-of-the-art results with our ResNet50-ensemble. In particular we outperform the strongest ensemble methods, including DREML~\cite{dreml} which utilize a much higher total embedding dimensionality. The large improvement is explained by the diverse and mutually complementary learning signals contributed by each task in our ensemble. In contrast, previous ensemble methods rely on the same, purely class-discriminative training signal for each learner. Note that some approaches strongly differ from the standard training protocols and architectures, resulting in more parameters and much higher GPU memory consumption, such as Rank~\cite{rankedlist} (32GB), ABE~\cite{abe} (24GB), Softtriple~\cite{softriple} and HORDE~\cite{horde}. Additionally, Rank~\cite{rankedlist} employs much larger batch-sizes to increase the number of classes per batch. This is especially crucial on the SOP dataset, which greatly benefits from higher class coverage due to its vast amount of classes~\cite{roth2020revisiting}. Nevertheless, our model outperforms these methods - in some cases even in its constrained version (IBN-512).

\subsection{Ablation Studies}
\begin{figure}[t]
    \centering
    \includegraphics[width=0.85\linewidth]{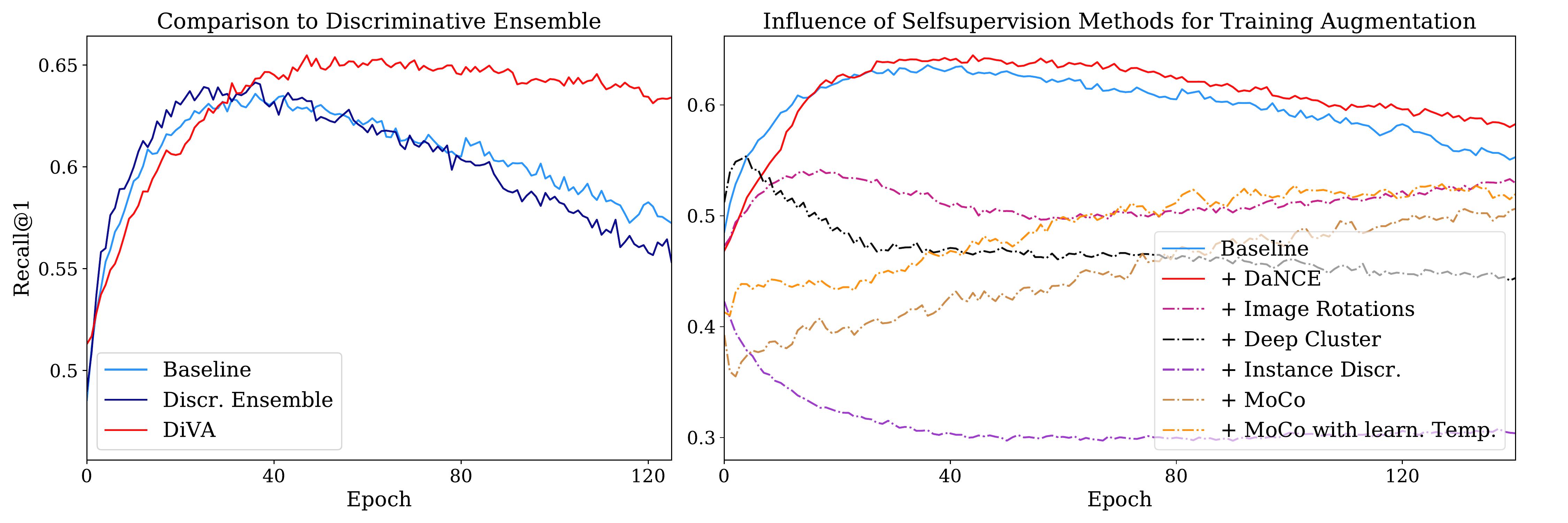}
    \caption{\textit{Analysis of complementary tasks for supervised learning.} (left): Performance comparison between class-dicsriminative training only (Baseline), ensemble of class-discriminative learners (Discr. Ensemble) and our proposed DiVA, which exhibits a large boost in performance. (right): Evaluation of self-supervised learning approaches combined with standard discriminative DML.}
    \label{fig:abl_ss_compare}
\end{figure}
\vspace{-4pt}
In this section we conduct ablation experiments for various parts of our model. For every ablation we again use the Inception-BN network. The dimensionality setting follows the performance study in sec.~\ref{sec:perf_study}. Again, we train each model with a fixed learning rate for fair comparison among ablations.
\\
\noindent
\textbf{Influence of distance-adaption in DaNCE.} To evaluate the benefit of our extension from $\mathcal{L}_{\text{nce}}$~\cite{moco,nce} to $\mathcal{L}_{\text{DaNCE}}$, we compare both versions in combination with standard supervised DML (i.e. class-discriminative features) in Fig.~\ref{fig:abl_ss_compare} (right). Our experiment indicates two positive effects: \textit{(i)} The training convergence with our extended objective is much faster and \textit{(ii)} the performance differs greatly between employing $\mathcal{L}_{\text{nce}}$ and $\mathcal{L}_{\text{DaNCE}}$. In fact, using the standard NCE objective is even detrimental to learning, while our extended version improves over the only discriminatively trained baseline. We attribute this to both the slow convergence of $\mathcal{L}_{\text{nce}}$ which is not able to support the faster discriminative learning and to emphasizing harder negatives in $\mathcal{L}_{\text{DaNCE}}$. In particular the latter is an important factor in ranking based DML~\cite{semihard}, as during training more and more negatives become uninformative. To tackle this issue, we also experimented with learning the temperature parameter $\tau$. While convergence speed increases slightly, we find no significant benefit in final generalization performance.
\\
\noindent
\textbf{Evaluation of self-supervision methods.} Fig.~\ref{fig:abl_ss_compare} (right) compares $\mathcal{\text{DaNCE}}$ to other methods from self-supervised representation learning. For that purpose we train the discriminative task with either DeepCluster~\cite{deepcluster}, RotNet~\cite{predict_rotations} or Instance Discrimination~\cite{wu2018unsupervised}. We observe that neither of these tasks is able to provide complementary information to improve generalization. DeepCluster, trained with 300 pseudo classes for classification, actually aims at approximating the class-discriminative learning signal while RotNet is strongly dependent on the variance of the training classes and converges very slowly. Instance discrimination seems to provide a contradictory training signal to the supervised task. These results are in line with previous works~\cite{hendrycksSSLAUG} which report difficulties to directly combine both supervised and self-supervised learning for improved test performance. In contrast, we explicitly adapt NCE to DML in our proposed objective DaNCE.
\begin{table*}[t]
    \setlength\tabcolsep{1.5pt}
    \centering
     \begin{tabular}{l|c|c|c|c}
        \toprule
        Methods $\rightarrow$  & Baseline & DiVA & No De-correlation & Separated models\\
        \hline
        Recall@1 $\rightarrow$ & 63.6     & 66.4 & 65.6              & 48.7\\        
        \bottomrule
    \end{tabular}
    \caption{\textit{Ablation studies}. We compare standard margin loss as baseline and DiVA performance against ablations of our model: no decorrelation between embeddings (No-Decorrelation.) and training an independent model for each task (Separated models). Total embedding dimensionality is 512.}
    \vspace{-1pt}
    \label{tab:ablations}
\end{table*}

\begin{SCfigure}
    \caption{\textit{Singular Value Spectrum.} We analyze the singular value spectrum of DiVA embeddings and that of a network trained with the standard discriminative task. We find that our gains in generalization performance (Tab.~\ref{tab:sota}, \ref{tab:baselines}) are reflected by a reduced spectral decay~\cite{roth2020revisiting} for our learned embedding space.}
    \includegraphics[width=0.4\linewidth]{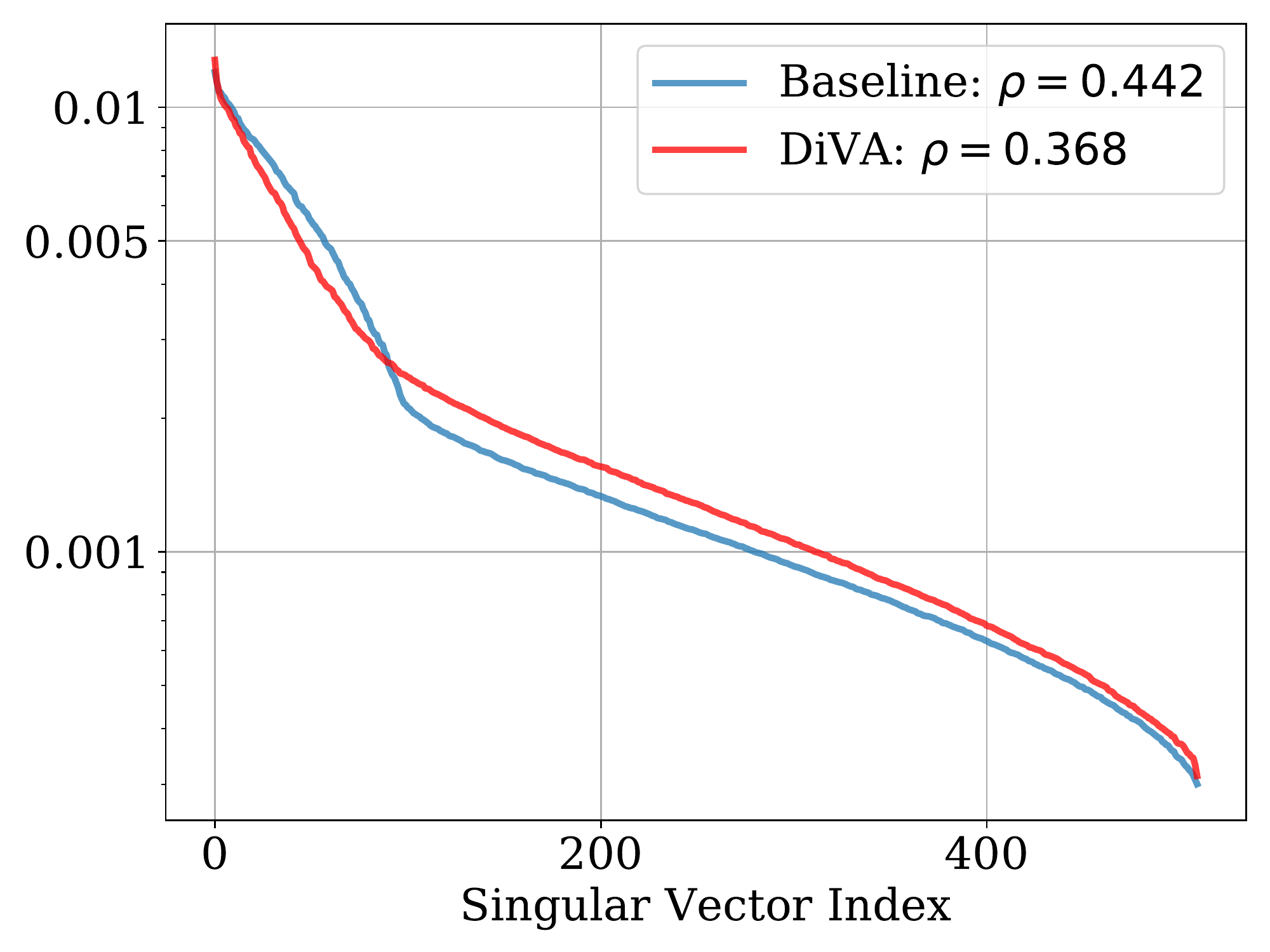}
    \label{fig:sv_spec}
\end{SCfigure}

\noindent
\textbf{Comparison to purely class-discriminative ensemble.} 
We now compare DiVA to an ensemble of class-discriminative learner (Discr. Ensemble) based on the same model architecture using embedding decorrelation in Fig.~\ref{fig:abl_ss_compare} (left). While the discriminative ensemble improves over the baseline, the amount of captured data information eventually saturates and, thus, performs significantly worse compared to our multi-feature DiVA ensemble. Further, our ablation reveals that joint optimization of diverse learning tasks regularizes training and reduces overfitting effects which eventually occur during later stages of DML training.
\\
\noindent
\textbf{Benefit of task decorrelation.} 
The role of decorrelating the embedding representations of each task during learning is analyzed by comparison to a model trained without this constraint. Firstly, Tab.~\ref{tab:ablations} demonstrates that omitting the decorrelation still outperforms the standard margin loss ('Baseline') by $2.1\%$ while operating on the same total embedding dimensionality. This proves that learning diverse features significantly improves generalization. Adding the de-corralation constraint then additionally boosts performance by $1.2\%$, as now each task is further encouraged to capture distinct data characteristics. 
\\
\noindent
\textbf{Learning without feature sharing.} To highlight the importance of feature sharing among our learning tasks, we train an individual, independent model for the class-discriminative, class-shared, sample-specific and intra-class task. At testing time, we combine their embeddings similar to our proposed model. Tab.~\ref{tab:ablations} shows a dramatic drop in performance to $48\%$ for the disconnected ensemble ('Separately Trained'), proving that sharing the complementary information captured from different data characteristics is crucial and mutually benefits learning. Without the class-discriminative signal, the other tasks lack the concept of an object class, which hurts the aggregation of embeddings (cf. Sec.~\ref{sec:Method}). 
\\
\noindent
\textbf{Generalization and embedding space compression.} 
Recent work~\cite{roth2020revisiting} links DML generalization to a decreased compression~\cite{tishby2015deep} of the embedding space. Their findings report that the number of directions with significant variance~\cite{manifoldmixup,roth2020revisiting} of a representation correlates with the generalization ability in DML. To this end, we analyze our model using their proposed spectral decay $\rho$ (lower is better) which is computed as the KL-divergence between the normalized singular value spectrum and a uniform distribution. Fig.~\ref{fig:sv_spec} compares the spectral decays of our model and a standard supervised baseline model. As expected, due to the diverse information captured, our model learns a more complex representation which results in a significantly lower value of $\rho$ and better generalization.

\section{Conclusion}
\label{sec:Conclusion}
In this paper we propose several learning tasks which complement the class-discriminative training signal of standard, supervised Deep Metric Learning (DML) for improved generalization to unknown test distributions. Each of our tasks is designed to capture different characteristics of the training data: class-discriminative, class-shared, intra-class and sample-specific features. For the latter, we adapt contrastive self-supervised learning to the needs of supervised DML. Jointly optimizing all tasks results in a diverse overall training signal which is further amplified by mutual decorrelation between the individual tasks. Unifying these distinct representations greatly boosts generalization over purely discriminatively trained models. 
\\
\\
\textbf{Acknowledgements} 
This work has been supported by hardware donations from NVIDIA (DGX-1), resources from Compute Canada, in part by Bayer AG and the German federal ministry BMWi within the project “KI Absicherung”.

\clearpage
%
%
\bibliographystyle{splncs04}
\bibliography{egbib}
\end{document}